\documentclass[10pt,twocolumn,letterpaper]{article}

\usepackage{iccv}
\usepackage{times}
\usepackage{epsfig}
\usepackage{graphicx}
\usepackage{amsmath}
\usepackage{amssymb}
\usepackage{multirow}
\usepackage{framed}
\usepackage[breaklinks=true,bookmarks=false]{hyperref}

\usepackage{caption, subcaption}

\usepackage[breaklinks=true,bookmarks=false]{hyperref}

\iccvfinalcopy 

\def\iccvPaperID{388} 

\begin{document}

\title{
       Learning Deep Representations Using Convolutional Auto-encoders with Symmetric Skip Connections
    }

\author{
    {
        \bf Jianfeng Dong$^\dag  $, Xiao-Jiao Mao$^\dag  $,  Chunhua Shen$^\star  $,  Yu-Bin Yang$^\dag  $
    }
        \\
        $^\dag  $State Key Laboratory for Novel Software Technology, Nanjing University, China\\
        $^\star $School of Computer Science, University of Adelaide, Australia
    }

\maketitle

\begin{abstract}


Unsupervised pre-training was a critical technique for training deep neural networks years ago.
With sufficient labeled data and modern training techniques, it is possible to train very deep neural networks from scratch in a purely supervised manner nowadays.
However, unlabeled data is easier to obtain and usually of very large scale.
How to make use of them better to help supervised learning is still a well-valued topic.
In this paper, 
we investigate convolutional denoising auto-encoders to show that unsupervised pre-training can still improve the performance of high-level image related tasks such as image classification and semantic segmentation.  
The architecture we use is a convolutional auto-encoder network with symmetric shortcut connections.
We empirically show that symmetric shortcut connections are very important for learning abstract representations via image reconstruction.
When no extra unlabeled data are available, unsupervised pre-training with our network can regularize the supervised training and therefore lead to better generalization performance.
With the help of unsupervised pre-training, our method achieves very competitive results in image classification using very simple all-convolution networks.
When labeled data are limited but extra unlabeled data are available, our method achieves good results in several semi-supervised learning tasks.

\end{abstract}

\section{Introduction}



Pre-training neural networks in a layer-wise unsupervised manner was shown to be able to ease the training procedure and improve the generalization performance\cite{dbn, learning_deep_architecture}.
Denoising auto-encoder\cite{denoising_auto_encoder} is one of the early methods for unsupervised pre-training, which can learn good representations via reconstruction. Unsupervised pre-training was regarded as a critical way to train deep neural networks at that time. As non-saturate activations\cite{alexnet, msra_init}, properer initialization\cite{xavier, msra_init} and sufficiently large labeled data been successfully used, it is possible to train very deep neural networks (more than 10 layers) in a purely supervised way\cite{vggnet, inception, resnet}, which was unimaginable years ago. It raises a question: can unsupervised pre-training still help to better train the modern deep architectures?

Suppose that a network is pre-trained on unlabeled dataset $A$ and fine-tuned on labeled dataset $B$. Based on the usage of training data, the unsupervised pre-training can be categorized into two cases:

(1) Case 1: ${A \subseteq B}$. As a data-driven network initialization method.

(2) Case 2: ${A \not\subseteq B}$. As a special type of semi-supervised learning method.

For case 1, Erhan \etal\cite{why_pre_training_help} showed that unsupervised pre-training with denoising auto-encoders improved generalization by ``initialization as regularization''. 
Recently, existing methods, especially those with convolutional neural networks\cite{context, video, inpainting}, mainly focus on Case 2 in order to employ unsupervised pre-training as a way for semi-supervised learning.

In this paper, we investigate unsupervised pre-training in the context of convolutional neural networks. The proposed network is essentially a stack of convolutional auto-encoders with symmetric shortcut connections. 
This network has recently been successfully used for image restoration tasks by Mao \etal\cite{rednet} and shows promising results. We show that using image reconstruction as unsupervised pre-training, this architecture learns good representations proper for high level image understanding tasks such as image classification and semantic segmentation.

When using our architecture in case 1, unsupervised pre-training without using extra unlabeled data can still regularize the network and improve generalization. In case 2 when extra unlabeled data are available, unsupervised pre-training with our architecture is also promising for semi-supervised learning.

\section{Related work}
\label{sec:related_work}

\textbf{Unsupervised Pre-training}
Using unsupervised pre-training to initialize neural networks and fine-tuning them in a supervised task can be dated back to deep belief network (DBN)\cite{dbn} and stacked auto-encoders\cite{learning_deep_architecture}. To prevent the auto-encoders from merely copying inputs during training, denoising auto-encoders \cite{denoising_auto_encoder} were proposed to learn representations from corrupted data. 
This greedy layer-wise pre-training in an unsupervised manner has been proved to be able to improve generalization performance. Erhan \etal\cite{why_pre_training_help} showed that unsupervised pre-training acted as a spacial type of regularization.

Although, in recent years unsupervised pre-training is not as popular as it was for training convolutional neural networks, some works\cite{context, video, inpainting} showed that unsupervised pre-training can still help supervised training by learning good representation from extra unlabeled data. As the using of reconstruction-based objective for stacked denoising auto-encoders, these methods use some prediction tasks as weak supervisions. Doersch \etal\cite{context} proposed to learn representations by predicting the relevant positions of patches extracted from the image. Wang \etal \cite{video} used visual tracking to learn representations in an unsupervised manner from unlabeled videos. A more relevent work to ours is Context Encoder \cite{inpainting}. They learn features by inpainting images with zero-masking corruption with a convolutional auto-encoder network. Although we also investigate masking noise, our network architecture can learn representations from any image reconstruction tasks besides inpainting. The usage of shortcut connections is annother difference. We will show experimentally that the shortcut connections are very important for unsupervised pre-training.

\textbf{Shortcut connections}
Our network architecture with shortcut connections mainly follows the work of image restoration by Mao \etal \cite{rednet}. They achieved promising low-level image restoration results by using the symmetrically connected encoder-decoder networks. The representations learned by this network show promises for low level restoration, while it is not clear whether they are also suitable for high-level tasks such as image classification. Thus, we investigate this problem in this paper. We will discuss more about the relationship between our method and the work of \cite{rednet} in Section \ref{sec:relation}.

As the networks go deeper, shortcut connections are introduced to tackle the problem of gradient vanishing/exploding during the training of deep residual networks\cite{resnet, new_resnet, wide_resnet}.
The shortcut connections, which learn "residual" functions, are linked from the input layer all the way up to the output layer.
For auto-encoder-based networks, symmetrically passing information from encoder to decoder has already been used in several tasks. Ranzato \etal\cite{shortcut1} passed transformation parameters from encoder to decoder to help the encoder to learn invariance unsupervised features about ``what'' without having to preserve local details about ``where''. Recently, similar auto-encoder architectures are also used in stacked what-where auto-encoders (SWWAE) for semi-supervised learning and SegNet for semantic segmentation. However, these methods mainly passed pooling indices to decoder forwardly. While in our method, the shortcut connections pass feature maps to decoder and also help to back-propagate gradients during training.

Another relevant work using similar architecture is Ladder network\cite{ladder_network, ladder_semi}, which is a special category of denoising auto-encoders for unsupervised and semi-supervised learning. It adds denoising objectives to all the hidden layers, and also engages lateral connections from encoder to decoder to discard local details of input data. The network we use can also be seen as a simpler version of Ladder Network: (1) We use a single overall objective instead of per-layer losses, (2) our shortcut connections employ element-wise addition forwardly while Ladder Network uses a pre-defined combinator function with extra parameters, and (3) we focus on investigating unsupervised pre-training for deep architectures while the impressive results of Ladder Network are mainly in a jointly semi-supervised manner\cite{ladder_semi}.

\section{Symmetrically Connected Convolutional Auto-encoders}
\label{sec:method}

\subsection{Architecture}

\begin{figure}[ht]
    \centering
    
    \begin{subfigure}[]{0.5\linewidth}
        \includegraphics[width=\linewidth]{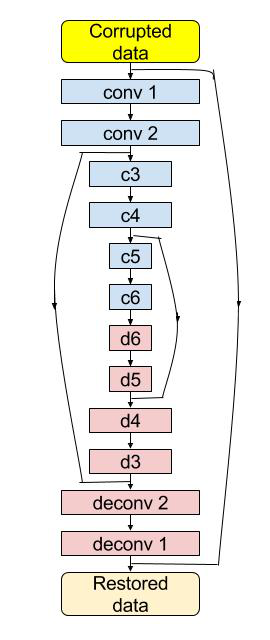}
        \caption{Network architecture}
        \label{fig:architecture_a}
    \end{subfigure}%
    \begin{subfigure}[]{0.5\linewidth}
        \includegraphics[width=\linewidth]{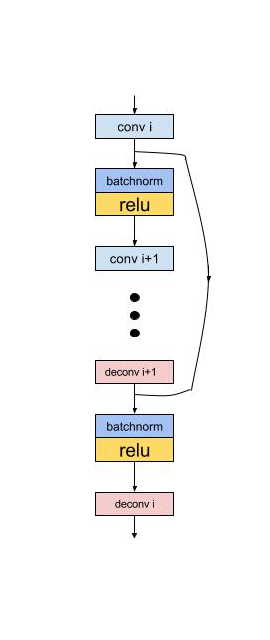}
        \caption{Shortcut connection}
        \label{fig:architecture_b}
    \end{subfigure}%
    
    \caption{Network architecture and the topology of a single shortcut connection.}
    \label{fig:architecture}
\end{figure}

The basic architecture of our network is a fully convolutional auto-encoder.
The encoder part is a chain of convolutional layers,
and the decoder part is a chain of decovolutional layers symmetric to the convolutional ones.
The corresponding encoder and decoder layers are connected by shortcut connections.
Each convolutional/deconvolutional layer is followed by a Batch Normalization \cite{batchnorm}
layer and a ReLU non-linearity layer. An illustration is shown in Figure \ref{fig:architecture}.

\textbf{Encoder}
The encoder acts as feature extractor. 
We mainly use 3$\times$3 convolutional layers following VGGNet\cite{vggnet}.
As in \cite{all_conv}, down-sampling is conducted by convolution with stride 2 instead of pooling,
which can be harmful to image restoration tasks as argued in \cite{rednet}.
Convolutional neural networks without pooling have already been successfully used by several works on image classification\cite{all_conv, resnet, new_resnet}.
Note that although our encoder is simple, 
one could always incorporate more complex architectures such as \cite{alexnet,vggnet,inception}.
The only modification is constructing a suitable decoder. 
We choose a concise one because we mainly focus on the investigation of unsupervised pre-training instead of the influence of different network architectures.

\textbf{Decoder}
The decoder takes features learned by the encoder and reconstructs the ``clean'' images.
We use deconvolution as our decoder unit, which is often referred as learnable up-sampling operation in 
tasks such as semantic segmentation\cite{segmentation_deconv} and image generation\cite{dcgan}. 
Since we use convolution with strides as learnable down-sampling operation in encoder,
it is important to also make the up-sampling learnable.
The layers in the decoder are organized symmetrically to the ones in the encoder.
For a specific deconvolutinal layer, 
the size of its input/output is equal to the output/input of its corresponding convolutional layer.
The reason is to achieve pixel-wise correspondence, which will be described later, with the use of shortcut connections.

\textbf{Shortcut Connections}
In our network, shortcut connections are used to pass feature maps forwardly.
The feature maps from a shortcut connection and the connected deconvolutional layer are then added element-wise.
For a single shortcut connection, the connecting strategy follows the pre-activation version of deep residual network\cite{new_resnet} as shown in Figure \ref{fig:architecture_b}. But at the network scale, instead of block-by-block as in ResNet, our shortcut connections are linked symmetrically as shown in Figure \ref{fig:architecture_a}. 


\subsection{Corruption type}
\label{sec:corruption_type}


We investigate two types of corruptions for training .

\textbf{Pixel-level Gaussian noise.} 
It is a typical pixel-level corruption which is used by
 many image denoising works.
Given an image $x$, we add random Gaussian noise with 0 mean and standard deviation $\sigma$ 
to each pixel uniformly. $\sigma$ is also called noise level.

\textbf{Block masking noise.}
Another type of corruption is to set some pixels of a image to 0.
It is also used in vanilla denoising auto-encoder\cite{denoising_auto_encoder}.
We use a special case of the masking noise, in which the pixels dropped out are adjacent.
Images with this type of corruption are used by Pathak \etal\cite{inpainting} for their Context Encoder, and the derived reconstruction task is also called image inpainting. 

\begin{figure}[ht]
    \begin{subfigure}[normal]{0.5\linewidth}
        \centering
        \includegraphics[scale=0.25]{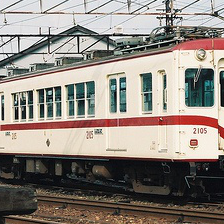}
    \end{subfigure}%
    \begin{subfigure}[normal]{0.5\linewidth}
        \centering
        \includegraphics[scale=0.25]{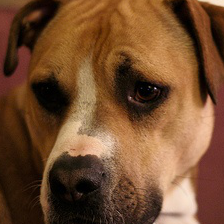}
    \end{subfigure}%
    
    \begin{subfigure}[t]{0.5\linewidth}
        \includegraphics[scale=0.25]{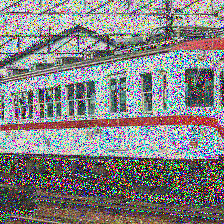}
        \includegraphics[scale=0.25]{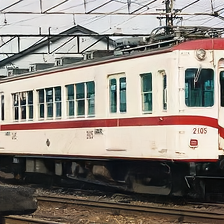}
    \end{subfigure}%
    \begin{subfigure}[t]{0.5\linewidth}
        \includegraphics[scale=0.25]{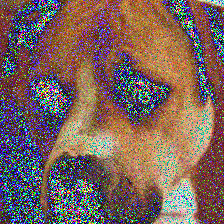}
        \includegraphics[scale=0.25]{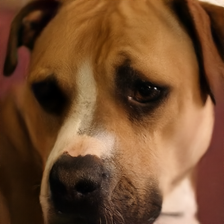}
    \end{subfigure}%
    
    
    \begin{subfigure}[t]{0.5\linewidth}
        \includegraphics[scale=0.25]{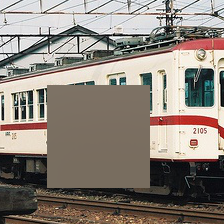}
        \includegraphics[scale=0.25]{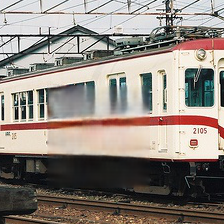}
    \end{subfigure}%
    \begin{subfigure}[t]{0.5\linewidth}
        \includegraphics[scale=0.25]{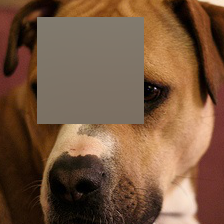}
        \includegraphics[scale=0.25]{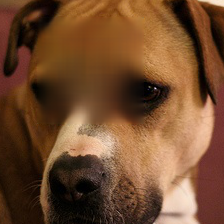}
    \end{subfigure}%
    
    \begin{subfigure}[t]{0.5\linewidth}
        \includegraphics[scale=0.25]{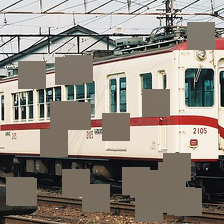}
        \includegraphics[scale=0.25]{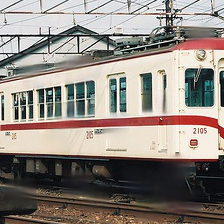}
    \end{subfigure}%
    \begin{subfigure}[t]{0.5\linewidth}
        \includegraphics[scale=0.25]{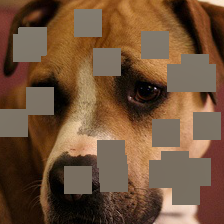}
        \includegraphics[scale=0.25]{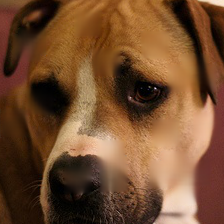}
    \end{subfigure}%
    
    \caption{Reconstruction performance on held-out images. 
    The top row indicates clean images. The following rows are 3 types of corrupted images and the reconstructed clean ones. From top to bottom: pixel Gaussian, masking noise with big block, and masking noise with small blocks.}
    \label{fig:reconstruction}
\end{figure}

We show the corrupted images as well as the reconstructions obtained by our network in Figure \ref{fig:reconstruction}. We find that using pixel-level Gaussian noise works better when the images are small and labeled data are sufficient. 
In other cases, masking noise with multiple small blocks always wins.

\subsection{Training Pipeline}
\label{sec:training_pipeline}

We follow the unsupervised pre-training and fine-tuning pipeline to learn representations from unlabeled data and then transfer them to new tasks. In contrast to the vanilla stacked denoising auto-encoders\cite{denoising_auto_encoder}, the pre-training is conducted end-to-end instead of greedy layer-wise.
During pre-training, for a clean image $x$, the input of the network is its corrupted version $\tilde{x}$. 
The output is the reconstruction represented as $f(\tilde{x})$. 
We train the network end-to-end by minimizing the mean square error between $x$ and $f(\tilde{x})$.
Although for masking noise, prior works\cite{denoising_auto_encoder, inpainting} also use a masked loss to emphasize the dropped pixels, we find that it is unnecessary when we use input-output shortcut connection to make the learned function as a residual one.
After training of the denoising network, 
we make use of the learned representations by fine-tuning the pre-trained layers.

\subsection{Unsupervised Pre-training as Initialization or Semi-supervised Learning}
\label{sec:init_and_semi}

For input data $X$ and its label $Y$, unsupervised pre-training the network essentially models the data distribution $P(X)$. But for most supervised fine-tuning tasks, we usually aim to model the conditional distribution $P(Y|X)$. Suppose that we pre-train the network on unlabeled dataset $A$ and fine-tune it on dataset $B$.

When ${A \subseteq B}$, the pre-training process does not have access to any extra data compared to the fine-tuning process. In this case, as shown by Erhan \etal\cite{why_pre_training_help}, the unsupervised pre-training improves generalization by regularizing the network through initialization. In contrast to some other initialization techniques\cite{xavier, msra_init}, the initialization is directly learned from training data.

when ${A \not\subseteq B}$, the unsupervised pre-training models $P(X)$ on A and transfer the learned representations to model $P(Y|X)$ on B. In this case, the learning pipeline can also be regarded as a way of semi-supervised learning since extra unlabeled data is used to help the learning from labeled data. Unlike some jointly semi-supervised learning methods\cite{ladder_network, swwae, catgan} which model $P(X)$ and $P(Y|X)$ simultaneously, unsupervised pre-training and fine-tuning model them separately. Hence during fine-tuning, the network can only access the information of unlabeled data through its initialization, which makes unsupervised pre-training not a perfect way for semi-supervised learning\cite{ladder_network, swwae}.

Although in recent years most works about unsupervised pre-training are conducted in semi-supervised setting \cite{context, video, inpainting}, we believe it is still important to figure out whether it can regularize the network via initialization. Hence we investigate both cases.

\subsection{Relationship to Image Restoration}
\label{sec:relation}

Since our network architecture is mainly inspired by the work of Mao \etal\cite{rednet} for image restoration, it is important to point out the different concerns for low-level image restoration and unsupervised pre-training with auto-encoders. Although the basic assumption for denoising auto-encoders \cite{denoising_auto_encoder} is ``a good representation is one that can be obtained robustly from a corrupted input and that will be useful for recovering the corresponding clean input'', it does not indicate all the learned  features, which are able to recover a clean image, are representative enough for high-level tasks. This is an important motivation of this paper. The representations we learned should not only achieve good restoration performance, but also be suitable for high-level tasks.

One important component that differs our paper from Mao \etal\cite{rednet} is the use of down-sampling. It was shown to be harmful for image restoration\cite{rednet}, but very important for high-level tasks such as image classification\cite{alexnet, resnet}. As for shortcut connections, Mao \etal\cite{rednet} use it mainly to 1) pass rich low-level image details to the decoder, and 2) back-propagate gradients for easier network training. However, for unsupervised pre-training, the most important thing is how to learn abstract  representations and make the learned model $P(X)$ to be helpful for learning $P(Y|X)$. 
 More experiments and analysis are provided in Section 5.1 to show that network with shortcut connections can indeed learn abstract representations for high level image classification, instead of merely learn representations good for reconstruction.

\section{Implementation Details}
\label{sec:implementation}

Our method is implemented with MXNet\cite{mxnet}, Tensorflow\cite{tensorflow} and Keras\cite{keras}. ADAM\cite{adam} is used for optimization.
The learning rate is set as 0.0001 at first, and divided by 10 when the loss does not drop. If not specified, the weights are initialized by random Gaussian numbers with 0 mean and standard deviation 0.01.
Shortcut connections are linked every 2 convolutional layers to their corresponding deconvolutional layers, as well as one connection from the input to the output.

For pre-processing, all pixels in an image are zero-centered and normalized. We randomly horizontally flip images during training for all tasks except for CIFAR-10 with 4,000 labels. 
Since we use convolution with stride to perform down-sampling, in order to make the zero-padding sizes on each side equal, the images are cropped to a proper size (29$\times$29 for CIFAR-10 and CIFAR-100, 89$\times$89 for STL-10 and 225$\times$225 for PASCAL VOC). During training, the crops are taken randomly. At testing time, central crops are taken for CIFAR-10, CIFAR-100 and STL-10. For PASCAL VOC, we follow \cite{inpainting} to average the results of 10 random crops. 

As for corruptions, we use pixel Gaussian noise for CIFAR-10 and CIFAR-100 classification in Section \ref{sec:exp_cifar}, and masking noise with small blocks for all other tasks.
The corruption is added to the images at real time before being fed into the network.
More details about training and network architecture are listed in the supplementary material.

\section{Experiments}
\label{sec:experiment}

We firstly discuss the importance of using shortcut connections via ablation study and show that the representations learned by our network are indeed suitable for classification. Then we experimentally show that unsupervised pre-training with our convolutional auto-decoder network can improve generalization even without using extra unlabeled data. When there are more unlabeled data, pre-training also does well in the semi-supervised task and achieves competitive results compared to several jointly semi-supervised learning methods. Furthermore, we show our auto-encoder network can scale well to larger scale unlabeled dataset and deeper architectures. At last, to better understand the learned representations, we train the network with data of different distributions and visualize the learned representations.

\subsection{On the Importance of Shortcut Connections}

Since the element-wise symmetric shortcut connections are the most important component of our network, we firstly conduct experiments to show that pre-training with shortcut connections prepares the network initialization proper for the supervised training task.

\begin{figure}
\centering
\begin{subfigure}{0.7\linewidth}
  \centering
  \includegraphics[width=\linewidth]{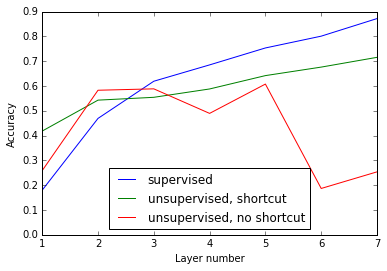}
  \caption{Classification}
  \label{fig:ab_classification}
\end{subfigure}%

\begin{subfigure}{0.7\linewidth}
  \centering
  \includegraphics[width=\linewidth]{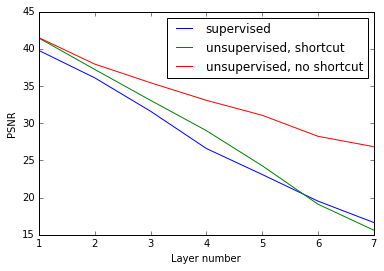}
  \caption{Reconstruction}
  \label{fig:ab_reconstruction}
\end{subfigure}
\caption{Comparisons: The classification and reconstruction performance of features extracted from each layers in different networks. Pre-training is conducted on Cifar-10.}
\label{fig:ablation}
\end{figure}

We train models on CIFAR-10 in 3 different settings:
1) an auto-encoder network with shortcuts, 2) an auto-encoder network without shortcuts \footnote{For all networks without shortcuts, we preserve shortcut from input to output, but discard all other internal shortcuts.} and 3) a classification network.
Then the learned parameters are fixed as feature extractor and transferred the features extracted from each encoder layer to two tasks: classification and reconstruction trained on 4,000 training images of CIFAR-10. For classification we only add a softmax regression on top of each ReLU activation. For reconstruction we use a simple stack of deconvolutional layers with the same numbers of parameters and different strides for activations of different sizes. The results are shown in Figure \ref{fig:ablation}.

One can clearly observe that as the network goes deeper, the features extracted from the layers of the supervised pre-trained network performs better on image classification but worse on image reconstruction. So do the features extracted from the layers of the pre-trained auto-encoder with shortcut connections. But the ones from the network without shortcut connections show a different tendency. Although the two auto-encoder networks both learn $P(X)$ instead of $P(Y|X)$, the network with shortcut connections tends to learn abstract representations by discarding image details, just as the classification network does. Hence our auto-encoder network with shortcut connections prepares the parameters properer for classification task. We believe this observation can also help explain the symmetric shortcut connections used by Ladder Network\cite{ladder_network}. Although they use a different shortcut strategy and train the network in a joint framework for semi-supervised learning.


For the auto-encoder network with shortcuts, the learned reconstructions from top layers' representations are poor as shown in Figure \ref{fig:ab_reconstruction}. However, we do observe that, during pre-training, its final reconstruction performance is better than that of the network without shortcuts, which is consistent to Mao \etal\cite{rednet}. This can be explained by learning good reconstruction also needs to learn good abstract representations, instead of only model richer local details of the corrupted data. In another word, abstract representations that help supervised learning can be learned by reconstruction, which is the core assumption of several methods\cite{inpainting, swwae, ladder_network} following denoising auto-encoders \cite{denoising_auto_encoder}.


\subsection{CIFAR-10 and CIFAR-100 Classification}
\label{sec:exp_cifar}

We evaluate unsupervised pre-training using our auto-encoder network with CIFAR-10 and CIFAR-100 classification, as a data-driven initialization technique and without using extra unlabeled data.
The architecture is a 15-layer all-convolution network. Detail configurations can be found in supplementary material. 

Table \ref{table:cifar_state} shows the overall classification accuracy 
of our network and other state-of-the-art results on CIFAR-10 and CIFAR-100.
One can observe that: 
1) Using a very simple network architecture, our method achieves competitive results by making use of the proposed unsupervised pre-training even without extra unlabeled data.
2) Shortcut connections during unsupervised pre-training can help to improve testing accuracy.
To our surprise, pre-training with auto-encoder network without using shortcuts does not perform so well. We believe it is due to the optimization problem during pre-training. The 15-layer encoder network leads to a 30-layer denoising network, which is difficult to optimize without using shortcut connections.

\begin{table}[]
\centering
\begin{tabular}{c|c|c}
Method                                              & CIFAR-10      & CIFAR-100     \\ \hline\hline
ELU \cite{elu}                                      & 93.45              & 75.72              \\
LSUV \cite{scalable_bayes}                          & 94.16              & 72.34              \\
ResNet 164 \cite{resnet}                            & 93.39              & 74.84              \\
ResNet V2 1001 \cite{new_resnet}                    & 95.08              & 77.29              \\
Wide ResNet \cite{wide_resnet}                      & \textbf{96.11}              & \textbf{81.15}              \\ \hline
Ours random Gaussian                                & 93.95              & 74.11              \\
Pre-training no shortcut                            & 92.07              & 70.21              \\
Pre-training shortcut                               & 95.15              & 75.41             
\end{tabular}
\caption{Comparisons with state-of-the-arts: CIFAR-10 and CIFAR-100 classification classification accuracy (\%).}
\label{table:cifar_state}
\end{table}

Although our classification results are slightly worse than those achieved by some recent works on deep residual network (ResNet)\cite{new_resnet, wide_resnet}, one should note that ResNet uses carefully designed very deep architecture, while we use a simple 15-layer all-convolution network. There is potential to combine our unsupervised pre-training strategy with ResNet to achieve even better accuracy. In fact, we have already successfully combined the wide ResNet\cite{wide_resnet} with our auto-encoder network and achieved better reconstruction performance during pre-training, but unfortunately, we find some optimization issues during fine-tuning due to the ResNet's dependency on SGD and weight decay. We describe our attempts and provide some intermediate results in supplementary material and leave it to future work.

\subsection{Classification with Limited Labels}

We show that our network can make use of extra unlabeled data and help to learn from limited labeled data in this section. In this case, unsupervised pre-training can also be seen as a semi-supervised learning method as we discussed in Section \ref{sec:init_and_semi}. We show that pre-training with our network can achieve competitive results even compared to several jointly semi-supervised learning methods.

\subsubsection{CIFAR-10 with 4000 Labeled Images}
We train our convolutional auto-encoders on all the 50,000 images of CIFAR-10 in an unsupervised manner, then fine-tune the encoder on 4,000 labeled images. For fair comparisons with previous works\cite{catgan, ladder_network, improved_gan}, we use a no pooling version of the 9-layer network in \cite{all_conv}. More details for the architecture and training process can be found in the supplementary material.

\begin{table}[]
\centering
\begin{tabular}{c|c}
Method                                              & Accuracy (\%)      \\ \hline\hline
Supervised state-of-the-art \cite{ladder_semi}      & 76.67$\pm$0.61 \\ \hline
Ladder Network \cite{ladder_semi}                   & 79.60$\pm$0.47 \\
CatGAN \cite{catgan}                                & 80.42$\pm$0.58 \\
Improved GAN \cite{improved_gan}                    & \textbf{81.37$\pm$2.32} \\ \hline
Ours no pre-training                                & 70.89$\pm$0.30               \\
Ours pre-training without shortcut                  & 74.07$\pm$0.43               \\
Ours pre-training with shortcut                     & 80.22$\pm$0.37              
\end{tabular}
\caption{Test accurasy on CIFAR-10 with 4000 labels}
\label{table:cifar_400}
\end{table}

Classicication results are shown in Table \ref{table:cifar_400}. Since unsupervised pre-training is not a common choice for this task, all the 3 methods listed in Table \ref{table:cifar_400} are under an jointly semi-supervised training framework, yet our method still achieves competitive results.

\subsubsection{STL-10 Classification}
STL-10 is a dataset for unsupervised and semi-supervised feature learning.
It contains 10 folds of labeled images with 1,000 images in each fold and 0.1 million extra unlabeled images.
We follow the common test protocols to firstly train our auto-encoder network on 100,000 unlabeled images, then fine-tune it on 10 pre-defined folds separately and report the average and standard deviation of the testing accuracy.
We use no pooling version of the architecture in \cite{swwae} for fair comparisons. 
\begin{table}[htb!]
\centering

\begin{tabular}{c|c|c}
Framework                                                                                  & Method                                 & Accuracy (\%)  \\ \hline\hline
\multirow{3}{*}{\begin{tabular}[c]{@{}c@{}}Unsupervised\\ Feature\\ Learning\end{tabular}} & Target Coding \cite{target_coding}     & 73.2      \\ \cline{2-3}
                                                                                           & Huang \etal \cite{attribute}           & \textbf{76.8$\pm$0.3} \\ \cline{2-3}
                                                                                           & Exemplar-CNN  \cite{discremitive}      & 75.4$\pm$0.3 \\ \hline
\begin{tabular}[c]{@{}c@{}}Jointly \\ Semi-supervised \\ Learning\end{tabular}             & SWWAE \cite{swwae}                     & 74.3      \\ \hline
\multirow{3}{*}{\begin{tabular}[c]{@{}c@{}}Unsupervised \\ Pre-training\end{tabular}}      & No pre-training                        & 64.6$\pm$0.7 \\ \cline{2-3}
                                                                                           & Ours without Shortcut                  & 70.2$\pm$0.6 \\ \cline{2-3}
                                                                                           & Ours with Shortcut                     & 75.8$\pm$0.5
\end{tabular}
\caption{STL-10 classification accuracy.}
\label{table:stl10}
\end{table}

The results are shown in Table \ref{table:stl10}. We achieve better results compared to jointly semi-supervised learning method SWWAE\cite{swwae}, which shows that shortcut connections are very important. We also list some results of unsupervised feature learning methods in Table \ref{table:stl10}. Although it is not a proper comparison because of the totally different training and testing pipeline.

\subsection{Learning from Large-scale Unlabeled Data}

In this section we show that our auto-encoder network can learn good representations from large-scale unlabeled data, and the decoder can also be transferred to supervised tasks.
Specifically, we train our convolutional auto-encoders on ImageNet 2012 training data without using any labels and then fine-tune it on the PASCAL VOC 2007 classification and segmentation tasks respectively. 

\subsubsection{PASCAL VOC Classification}

After adding fully connected layers, the pre-trained encoder is fine-tuned on the PASCAL VOC 2007 classification dataset\cite{voc}, . 

Most of the prior unsupervised pre-training methods using Alexnet\cite{alexnet} as their backbone architecture\cite{inpainting, context, video}. Unfortunately, we find it is difficult to converge when adapting AlexNet into our architecture. The reasons are: 
(1) AlexNet contains several severe down-sampling layers at first, which is quite harmful to reconstruction, and
(2) The layers in Alexnet are carefully hand-crafted to ease the optimization.
While the shortcut connections we use break the original data flow and make it harder to optimize.
Moreover, we attempt to test our method with very deep convolutional neural network architectures. We instead use a network based on VGG-16 network\cite{vggnet} as base architecture. For a fair comparison, we also re-implement the Context-Encoder \cite{inpainting} with this architecture. The architecture details are listed in supplementary material.
The results are reported in Table \ref{table:voc_compare}.



\begin{table}[htb!]
\centering
\begin{tabular}{l|l}
Method                & mAP(\%) \\ \hline\hline
Random Gaussian       & 67.11   \\
Context Encoder       & 70.24        \\ \hline
Ours without shortcut & 69.38   \\
Ours shortcut         & \textbf{71.25}  
\end{tabular}
\caption{PASCAL VOC classification results.}
\label{table:voc_compare}
\end{table}

\subsubsection{Transferring both Encoder and Decoder to Semantic Segmentation}

In traditional transfer learning using auto-encoders, the decoder part is discarded during fine-tuning. In this section, we show that the decoder of our network is also transferable. To do this, we evaluate our network on PASCAL VOC 2012 segmentation task. Note that our aim here is not to achieve state-of-the-art results since the network is not particularly designed for image segmentation. More importantly, most state-of-the-art methods are pre-trained on large-scale labeled data, while our pre-training is with unlabeled data.

\begin{figure}[htb!]
\centering
\includegraphics[width=0.43\textwidth]{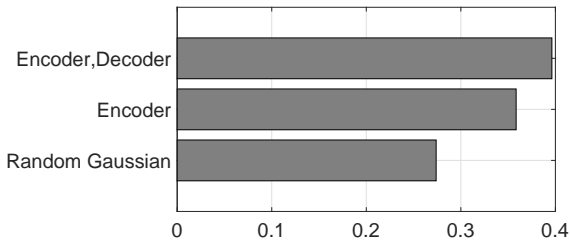}
\caption{Mean IUs of PASCAL VOC 2012 Segmentation task on validation data with different fine-tuning strategies.}
\label{table:segmentation}
\end{figure}

We can use our network to perform segmentation by simply replacing the last deconvolutional layer with a convolution layer of proper number of channels for segmentation.
3 segmentation networks are trained with different initialization strategies:
(1) initializing all layers with small random Gaussian numbers,
(2) initializing the encoder by unsupervised pre-training and initializing the decoder randomly,
and (3) initializing both the encoder and decoder by pre-training.
The mean intersection over union scores on validation images are reported in Figure \ref{table:segmentation}.

It is easy to observe that initializing decoder with pre-training leads to better segmentation results. Since most state-of-the-art CNN-based segmentation methods\cite{segmentation_deconv, fcn, segnet} using encoder-decoder architectures normally initialize the encoder by supervised pre-training and initialize the decoder randomly, this experiment also shows a potential to initialize the decoder with proper unsupervised pre-training to further improve the segmentation performance.

\subsection{Analysis on the Learned Representations}

\subsubsection{Training on Images of Different Distributions}
\label{sec:distribution}

The 20 classes in PASCAL VOC dataset can be categorized into 4 coarse classes:
animal, vehicle, indoor and person.
In this section, we investigate whether pre-training on images with different
distributions affects the testing performance after fine-tuning.
For instance, if the network sees a lot of animal images during pre-training,
we would like to know if it could better recognize animals.

Specifically, we take two subsets from Imagenet 2012 training images as our unlabeled datasets. Images of the first subset belong to super class ``conveyance, transport'', and those of the second belong to ``mammal'' or ``bird''. We select 93 thousand images from each subset for balance.
We use two coarse classes instead of fine-grained classes because the data for pre-training is assumed to be unlabeled. In practice, it is usually much easier to collect unlabeled data which belong to a coarse category such as ``animal'' or ``vehicle''.

The mean average precision score for each coarse class is shown in Table \ref{table:class_compare}. 
We can observe that the network pre-trained with vehicle images works better on recognizing vehicles, and the animal trained one works better on recognizing animals.
Meanwhile, in each class, two networks with unsupervised pre-training outperform the one trained from scratch.

\begin{table}[htb!]
\centering

\begin{tabular}{c|ccc}
\hline
Pre-trained on  & Animal  & Vechicle & Mean All \\ \hline\hline
No pre-training       & 67.02 & 76.87 & 67.11 \\
Animal 93K  & \textbf{73.04} & 79.04  & 70.77 \\
Vehicle 93K & 72.17 & \textbf{80.00}  & \textbf{70.93} \\ \hline
\end{tabular}
\caption{Compare pre-training with different data distributions: Mean average precision (\%) of PASCAL VOC 2007 classification on different coarse classes.}
\label{table:class_compare}
\end{table}

This indicates that our network learns semantic relevant representations during pre-training from unlabeled data of different distributions. We will further prove this with some visualizations in next section.

\subsubsection{Visualization}

\begin{figure}[ht]
    \begin{subfigure}[t]{0.33\linewidth}
        \includegraphics[scale=0.33]{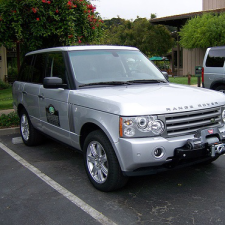}
        
        \includegraphics[scale=0.33]{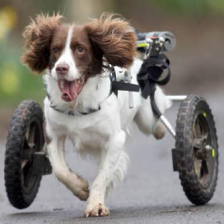}
        
        \includegraphics[scale=0.33]{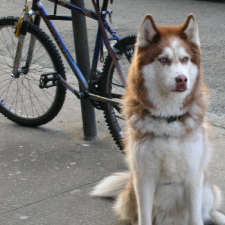}
        
        \caption{Original Images}
    \end{subfigure}%
    \begin{subfigure}[t]{0.33\linewidth}
        \includegraphics[scale=0.33]{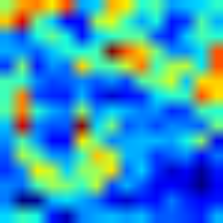}
        
        \includegraphics[scale=0.33]{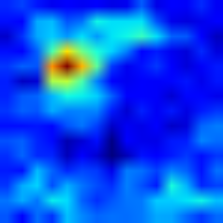}
        
        \includegraphics[scale=0.33]{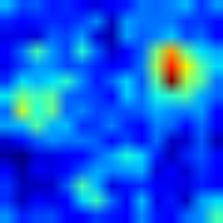}
        
        \caption{Animal Trained}
    \end{subfigure}%
    \begin{subfigure}[t]{0.33\linewidth}
        \includegraphics[scale=0.33]{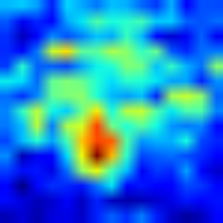}
        
        \includegraphics[scale=0.33]{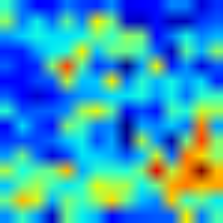}
        
        \includegraphics[scale=0.33]{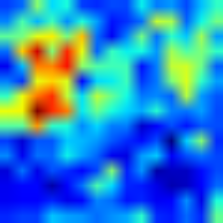}
        
        \caption{Vehicle Trained}
    \end{subfigure}%
    
    \caption{Visualization of feature maps obtained by the convolutional auto-encoders trained on data with different distributions. The network learns these merely by reconstructing corrupted images with different distributions, and no annotated label is used.}
    \label{fig:car_dog}
\end{figure}

We visualize the activations of the 12th ReLU layer of the network in Section \ref{sec:distribution}. The feature maps are directly obtained from the unsupervised pre-trained auto-encoders before fine-tuning. Visualizations are shown in Figure \ref{fig:car_dog}, in which all the images are not included in the training data.

\begin{figure}[ht]
    \begin{subfigure}[t]{\linewidth}
        \includegraphics[scale=0.2]{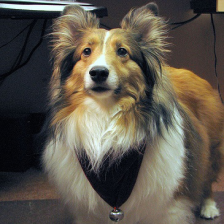}
        \includegraphics[scale=0.2]{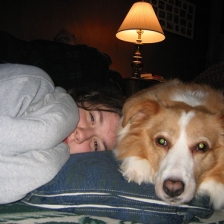}
        \includegraphics[scale=0.2]{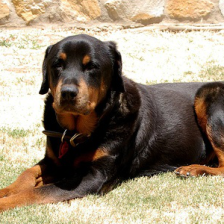}
        \includegraphics[scale=0.2]{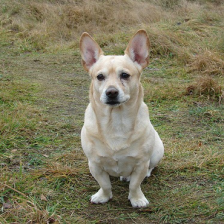}
        \includegraphics[scale=0.2]{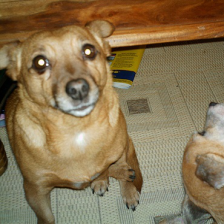}
        
        \includegraphics[scale=0.2]{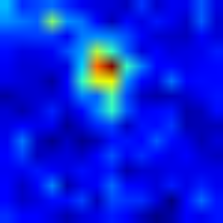}
        \includegraphics[scale=0.2]{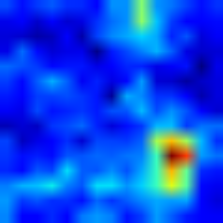}
        \includegraphics[scale=0.2]{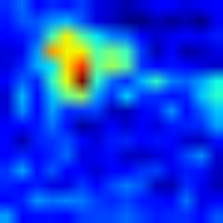}
        \includegraphics[scale=0.2]{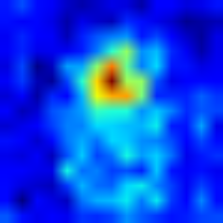}
        \includegraphics[scale=0.2]{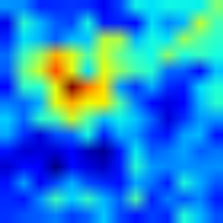}
        
        \caption{Clean Images and Features}
        \label{fig:detect_dog_a}
    \end{subfigure}%

    \begin{subfigure}[t]{\linewidth}
        \includegraphics[scale=0.2]{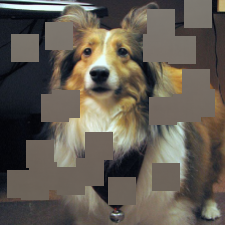}
        \includegraphics[scale=0.2]{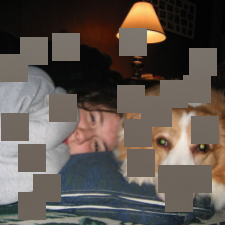}
        \includegraphics[scale=0.2]{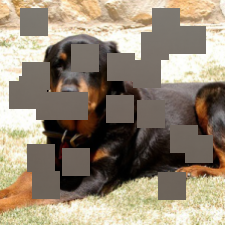}
        \includegraphics[scale=0.2]{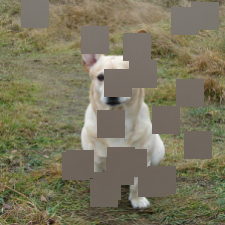}
        \includegraphics[scale=0.2]{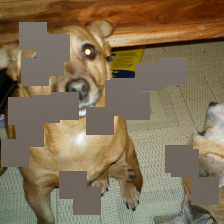}
        
        \includegraphics[scale=0.2]{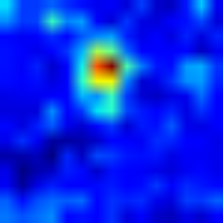}
        \includegraphics[scale=0.2]{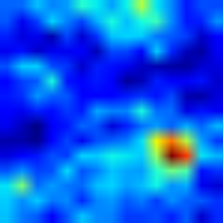}
        \includegraphics[scale=0.2]{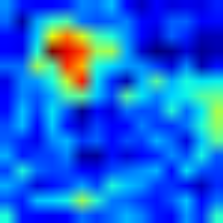}
        \includegraphics[scale=0.2]{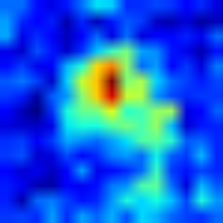}
        \includegraphics[scale=0.2]{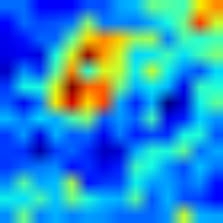}
        
        \caption{Corrupted Images and Features}
        \label{fig:detect_dog_b}
    \end{subfigure}%

    \caption{Visualization of more features of animal trained network. The features change little even when then dog faces are partially occluded.}
    \label{fig:detect_dog}
\end{figure}

The animal trained network shows more attention on dog faces,
while the vehicle trained network is more sensitive to wheels.
It is intriguing because the networks learn these characteristics only by reconstruction,
and there are no annotated labels indicating where dogs or cars are in the training images.
The reasons may be:
(1) ``dog'' and ``car'' are dominant classes in each training data,
hence the network ``remember'' some typical patterns during reconstruction, and 
(2) complex image contents like dog faces and wheels are usually harder to reconstruct than other parts,
hence the network gives them more attention.

We show more visualization of the animal trained auto-encoder in Figure \ref{fig:detect_dog_a}. 
It can act as a simple unsupervised trained dog face detector in these cases,
although the training images are not all dogs but with other kinds of animals.
We further show the feature maps of corrupted images in Figure \ref{fig:detect_dog_b}. 
One can observe that the learned representations change little even when the dog faces are partially occluded, 
which shows that the representations learned by our convolutional auto-encoders are very robust to corruptions.

\section{Conclusions}

We propose a method for unsupervised pre-training by image reconstruction. The network architecture we use is a fully convolutional encoder-decoder network with symmetric shortcut connections. We show  experimentally shortcut connections are critical to help auto-encoder network learn good representations during pre-training. 
With proposed unsupervised pre-training method, we achieve competitive results on classification tasks with a very simple all-convolution neural network. We also use unsupervised pre-training method in semi-supervised learning tasks and achieve competitive results compared to several jointly semi-supervised learning methods.


{\small
\bibliographystyle{ieee}
\bibliography{reference}
}

\clearpage
\begin{center}
\textbf{\large Supplemental Material}
\end{center}

\setcounter{section}{0}
\setcounter{figure}{0}
\setcounter{table}{0}
\setcounter{page}{1}
\renewcommand\thesection{\Alph{section}}

\section{Network Architecture}

The detailed network settings are shown in Table \ref{table:architecture}. Network-1 is used for CIFAR and STL-10. For CIFAR-10 and CIFAR-100, we choose $m_1=5, m_2=5, m_3=5, m_4=0, n=128$. For STL-10, the parameters are $m_1=2, m_2=2, m_3=3, m_4=3, n=64$, which lead to a similar architecture to the base network used by SWWAE\cite{swwae}. Network-2 is used for training on CIFAR-10 with 4,000 labels. It is based on the All-CNN-C model in \cite{all_conv} and similar architectures are used by the compared methods. For all experiments on PASCAL VOC, we use Network-3 which is based on VGG-16 network\cite{vggnet} but without pooling.

\begin{table}[htb!]
\centering
\begin{tabular}{c|c|c}

Network-1                                                               & Network-2                                                                    & Network-3                                                               \\ \hline
\begin{tabular}[c]{@{}c@{}}conv3x3,n\\ ...\\ $\times(m_1-1)$\end{tabular}      & \begin{tabular}[c]{@{}c@{}}conv3x3,96\\ conv3x3,96\end{tabular}              & conv3x3,64                                                              \\ \hline
conv3x3/2,n                                                             & conv3x3/2,96                                                                 & conv3x3/2,64                                                            \\ \hline
\begin{tabular}[c]{@{}c@{}}conv3x3,2n\\ ...\\ $\times(m_2-1)$\end{tabular}        & \begin{tabular}[c]{@{}c@{}}conv3x3,192\\ conv3x3,192\end{tabular}         & conv3x3,128                                                             \\ \hline
conv3x3/2,2n                                                            & conv3x3/2,192                                                                & conv3x3/2,128                                                           \\ \hline
\begin{tabular}[c]{@{}c@{}}conv3x3,4n\\ ...\\ $\times(m_3-1)$\end{tabular}        & \begin{tabular}[c]{@{}c@{}}conv3x3,192\\ conv1x1,192\end{tabular}            & \begin{tabular}[c]{@{}c@{}}conv3x3,256\\ conv3x3,256\end{tabular}       \\ \hline
conv3x3/2,4n                                                            &                                                                              & conv3x3/2,256                                                           \\ \hline
\begin{tabular}[c]{@{}c@{}}conv3x3,8n\\ ...\\ $\times(m_4-1)$\end{tabular}        &                                                                              & \begin{tabular}[c]{@{}c@{}}conv3x3,512\\ conv3x3,512\end{tabular}       \\ \hline
conv3x3/2,8n                                                            &                                                                              & conv3x3/2,512                                                           \\ \hline
                                                                        &                                                                              & \begin{tabular}[c]{@{}c@{}}conv3x3,512\\ conv3x3,512\end{tabular}       \\ \hline
                                                                        &                                                                              & conv3x3/2,512                                                           \\ \hline
\begin{tabular}[c]{@{}c@{}}global\_ave\_pool\\ fc,n\_class\end{tabular} & \begin{tabular}[c]{@{}c@{}}conv1x1,n\_class\\ global\_ave\_pool\end{tabular} & \begin{tabular}[c]{@{}c@{}}fc,4096\\ fc,4096\\ fc,n\_class\end{tabular}

\end{tabular}

\caption{Network architectures. }
\label{table:architecture}
\end{table}

\section{More Training Details}

We use ADAM as optimizer. The learning rate is set to 0.002 for training on CIFAR-10 with 4,000 labels and 0.0001 for other tasks. The denoising network is trained with 120 epochs on CIFAR-10 and CIFAR-100, 40 epochs on STL-10 and 6 epochs on Imagenet. The learning rate is multiplied by 0.1 at the 60th and 90th epoch for CIFAR, the 30th epoch for STL-10 and the 4th epoch for Imagenet. For supervised fine-tuning, we train the network with 200 epochs on CIFAR-10, CIFAR-100, STL-10 and 400 epochs on PASCAL VOC. The learning rate is multiplied by 0.2 at the 60th, 120th and 160th epoch for CIFAR-10, CIFAR-100 and STL-10 respectively. For PASCAL VOC the learning rate is multiplied by 0.1 at the 200th and 300th epoch. 

When the labeled data are limited (CIFAR-10 with 4,000 labels and STL-10), we find that it is better to firstly fix the pre-trained parameters and only fine-tune the randomly initialized layers for 10 epochs, and then fine-tune all layers. But it is not necessary for other tasks. And for the fine-tuning on CIFAR-10 with 
4,000 labels, adding the same corruption used in pre-training with 50 percent chance slightly improves the performance. But when training the network with random initialization, the performance degrades.

\section{Combining with ResNet}

Although one can combine the shortcut connections with the ones used by ResNet \cite{resnet}, we find it is difficult to obtain higher accuracy for ResNet even with unsupervised pre-training. The reasons may be:

ResNet uses stochastic gradient descent (SGD) with momentum for optimization. We find that compared to SGD, training ResNet with ADAM leads to faster convergence but far lower accuracy even with weight decay. This may be because SGD acts as a special type of regularization during training but using ADAM causes overfitting. For our reconstruction-based unsupervised pre-training, the story is different. When training on CIFAR-100 with $\sigma=30$ Gaussian noise, the final PSNR of ADAM trained network is higher than the SGD trained network by around 1.0dB, which is a huge difference for image reconstruction task. Moreover, we find it is important to use the same optimizer for pre-training and fine-tuning.

Weight decay is very important for ResNet to generalize well. But we find that the learned parameters result in 
quite small magnitude, which usually does not happen in supervised training, when training the image reconstruction network even with a small weight decay. A possible reason is that generative models are less likely to overfit. If weight decay is not used for pre-training but for fine-tuning, we observe lower accuracy even compared to randomly initialized network.

We believe that in order to improve the performance of ResNet by using unsupervised pre-training, it is important to firstly figure out how does SGD help to regularize ResNet and how does weight decay influence pre-training and fine-tuning.

\section{Reconstruction Performance}

\begin{figure*}[ht]
    \centering

    \begin{subfigure}[t]{\linewidth}
        \centering

        \includegraphics[scale=0.35]{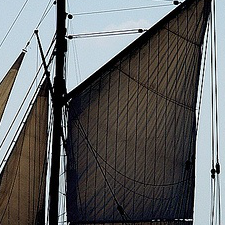}
        \includegraphics[scale=0.35]{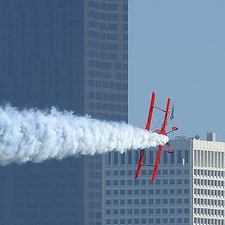}
        \includegraphics[scale=0.35]{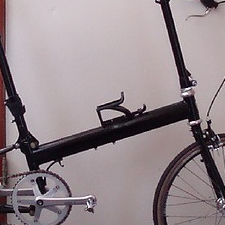}
        \includegraphics[scale=0.35]{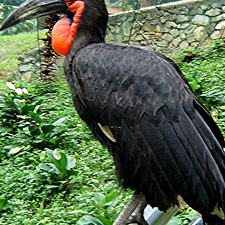}
        \includegraphics[scale=0.35]{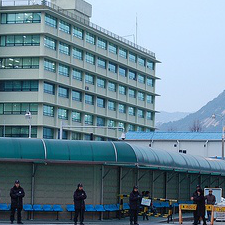}
        \includegraphics[scale=0.35]{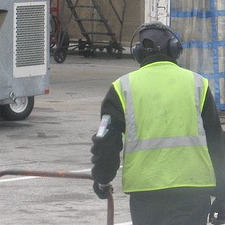}
        
    \end{subfigure}%
    
    \bigskip
    
    \begin{subfigure}[t]{\linewidth}
        \centering
    
        \includegraphics[scale=0.35]{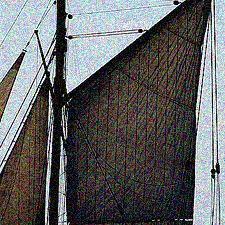}
        \includegraphics[scale=0.35]{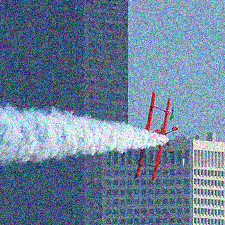}
        \includegraphics[scale=0.35]{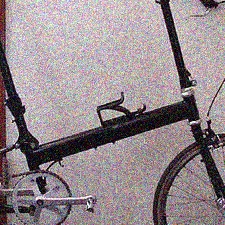}
        \includegraphics[scale=0.35]{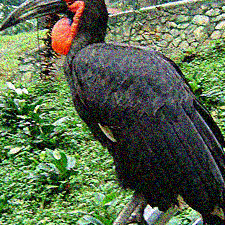}
        \includegraphics[scale=0.35]{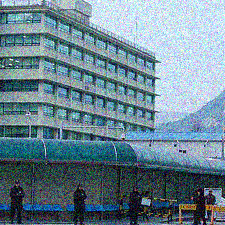}
        \includegraphics[scale=0.35]{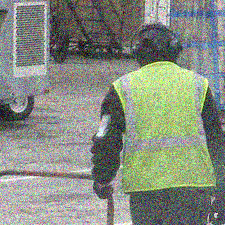}
        
        \includegraphics[scale=0.35]{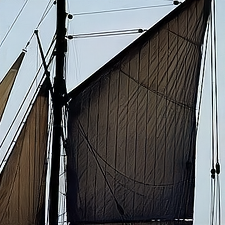}
        \includegraphics[scale=0.35]{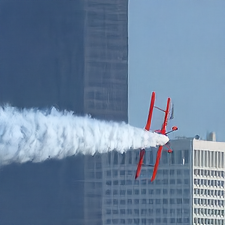}
        \includegraphics[scale=0.35]{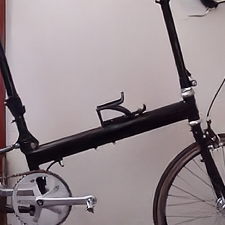}
        \includegraphics[scale=0.35]{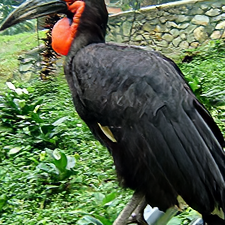}
        \includegraphics[scale=0.35]{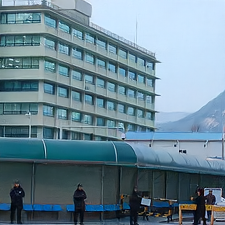}
        \includegraphics[scale=0.35]{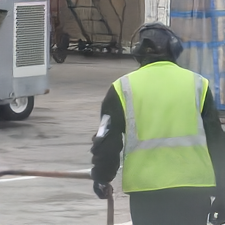}
    \end{subfigure}%
    
    \bigskip
    
    \begin{subfigure}[t]{\linewidth}
        \centering
    
        \includegraphics[scale=0.35]{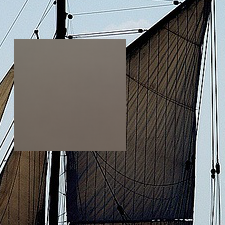}
        \includegraphics[scale=0.35]{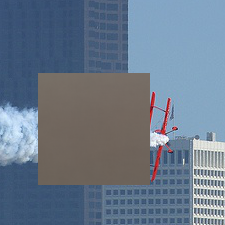}
        \includegraphics[scale=0.35]{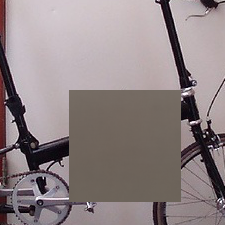}
        \includegraphics[scale=0.35]{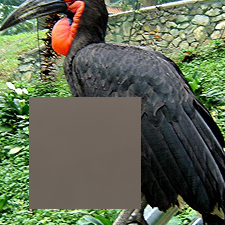}
        \includegraphics[scale=0.35]{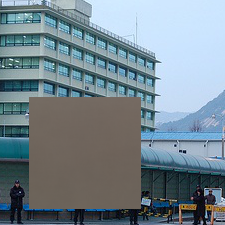}
        \includegraphics[scale=0.35]{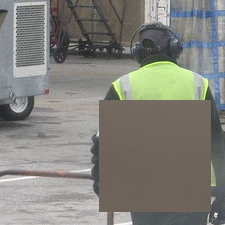}
        
        \includegraphics[scale=0.35]{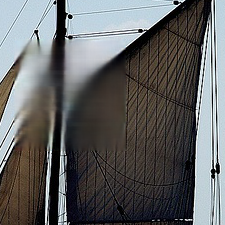}
        \includegraphics[scale=0.35]{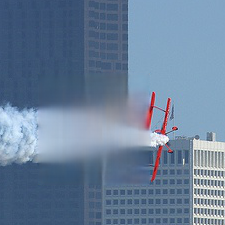}
        \includegraphics[scale=0.35]{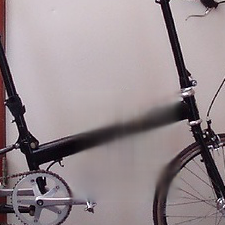}
        \includegraphics[scale=0.35]{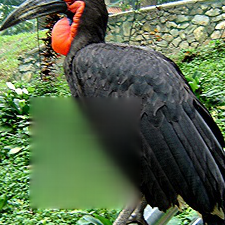}
        \includegraphics[scale=0.35]{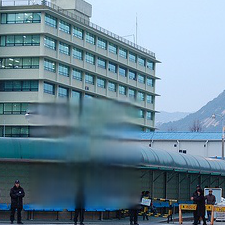}
        \includegraphics[scale=0.35]{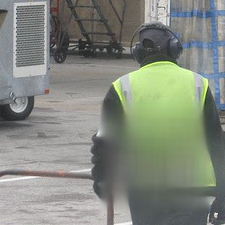}
    \end{subfigure}%
    
    \bigskip
    
    \begin{subfigure}[t]{\linewidth}
        \centering
    
        \includegraphics[scale=0.35]{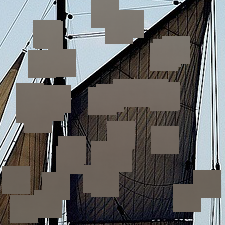}
        \includegraphics[scale=0.35]{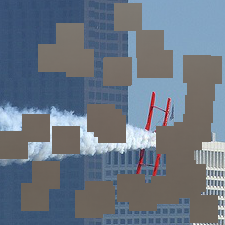}
        \includegraphics[scale=0.35]{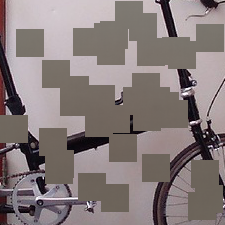}
        \includegraphics[scale=0.35]{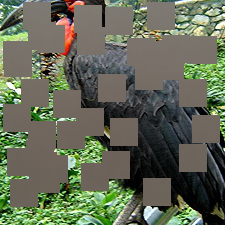}
        \includegraphics[scale=0.35]{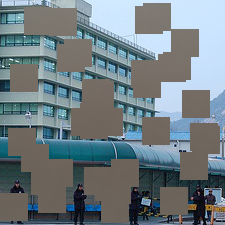}
        \includegraphics[scale=0.35]{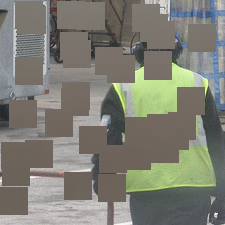}
        
        \includegraphics[scale=0.35]{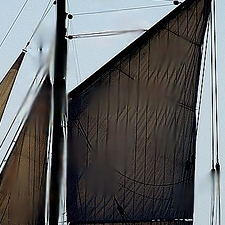}
        \includegraphics[scale=0.35]{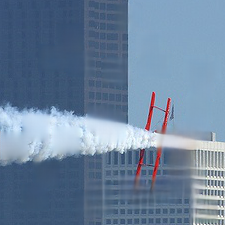}
        \includegraphics[scale=0.35]{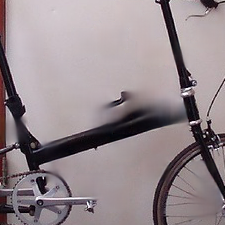}
        \includegraphics[scale=0.35]{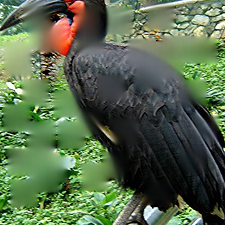}
        \includegraphics[scale=0.35]{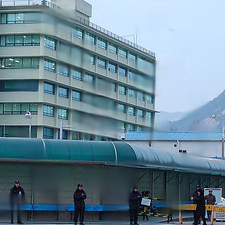}
        \includegraphics[scale=0.35]{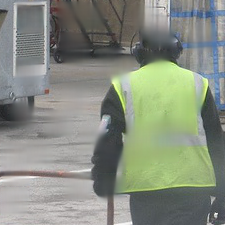}
    \end{subfigure}%

    \caption{Reconstruction on PASCAL VOC images. The training is conducted on training data of Imagenet 2012. Images at the top line are clean ones.}
    \label{fig:voc_vis}
\end{figure*}

\begin{figure*}[ht]
    \centering

    \newcommand\myscale{1.2}

    \begin{subfigure}[t]{\linewidth}
        \centering

        \includegraphics[scale=\myscale]{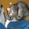}
        \includegraphics[scale=\myscale]{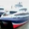}
        \includegraphics[scale=\myscale]{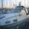}
        \includegraphics[scale=\myscale]{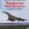}
        \includegraphics[scale=\myscale]{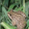}
        \includegraphics[scale=\myscale]{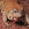}
        \includegraphics[scale=\myscale]{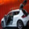}
        \includegraphics[scale=\myscale]{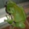}
        \includegraphics[scale=\myscale]{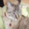}
        \includegraphics[scale=\myscale]{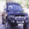}
        \includegraphics[scale=\myscale]{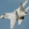}
        \includegraphics[scale=\myscale]{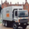}
        
    \end{subfigure}%
    
    \bigskip
    
    \begin{subfigure}[t]{\linewidth}
        \centering
    
        \includegraphics[scale=\myscale]{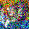}
        \includegraphics[scale=\myscale]{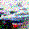}
        \includegraphics[scale=\myscale]{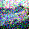}
        \includegraphics[scale=\myscale]{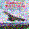}
        \includegraphics[scale=\myscale]{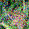}
        \includegraphics[scale=\myscale]{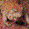}
        \includegraphics[scale=\myscale]{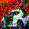}
        \includegraphics[scale=\myscale]{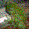}
        \includegraphics[scale=\myscale]{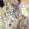}
        \includegraphics[scale=\myscale]{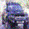}
        \includegraphics[scale=\myscale]{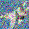}
        \includegraphics[scale=\myscale]{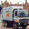}
        
        \includegraphics[scale=\myscale]{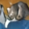}
        \includegraphics[scale=\myscale]{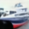}
        \includegraphics[scale=\myscale]{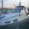}
        \includegraphics[scale=\myscale]{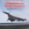}
        \includegraphics[scale=\myscale]{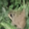}
        \includegraphics[scale=\myscale]{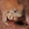}
        \includegraphics[scale=\myscale]{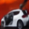}
        \includegraphics[scale=\myscale]{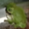}
        \includegraphics[scale=\myscale]{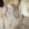}
        \includegraphics[scale=\myscale]{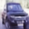}
        \includegraphics[scale=\myscale]{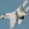}
        \includegraphics[scale=\myscale]{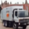}
    \end{subfigure}%

    \caption{Reconstruction on CIFAR-10 testing images. The training is conducted on training data of CIFAR-10. Images at the top line are clean ones.}
    \label{fig:cifar_10_vis}
\end{figure*}

\begin{figure*}[ht]
    \centering

    \newcommand\myscale{0.9}

    \begin{subfigure}[t]{\linewidth}
        \centering

        \includegraphics[scale=\myscale]{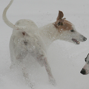}
        \includegraphics[scale=\myscale]{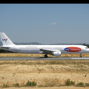}
        \includegraphics[scale=\myscale]{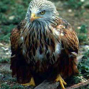}
        \includegraphics[scale=\myscale]{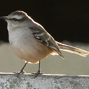}
        \includegraphics[scale=\myscale]{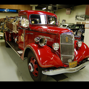}
        \includegraphics[scale=\myscale]{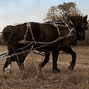}
        
    \end{subfigure}%
    
    \bigskip
    
    \begin{subfigure}[t]{\linewidth}
        \centering
    
        \includegraphics[scale=\myscale]{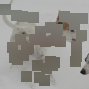}
        \includegraphics[scale=\myscale]{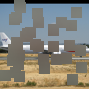}
        \includegraphics[scale=\myscale]{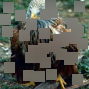}
        \includegraphics[scale=\myscale]{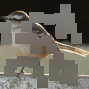}
        \includegraphics[scale=\myscale]{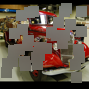}
        \includegraphics[scale=\myscale]{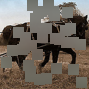}
        
        \includegraphics[scale=\myscale]{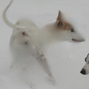}
        \includegraphics[scale=\myscale]{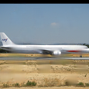}
        \includegraphics[scale=\myscale]{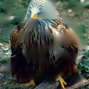}
        \includegraphics[scale=\myscale]{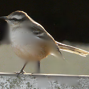}
        \includegraphics[scale=\myscale]{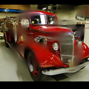}
        \includegraphics[scale=\myscale]{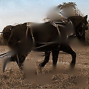}
    \end{subfigure}%

    \caption{Reconstruction on STL-10 testing images. The training is conducted on 100 thousand unlabeled images of STL-10. Images at the top line are clean ones.}
    \label{fig:stl_10_vis}
\end{figure*}

Although our aim is to learn better representations for high-level vision tasks instead of simply improving the image reconstruction performance, we do observe that as other parameters being fixed, a better image reconstruction performance always leads to a higher accuracy after fine-tuning. More reconstruction results on different datasets are shown in Figure \ref{fig:voc_vis}, Figure  \ref{fig:cifar_10_vis} and Figure \ref{fig:stl_10_vis}.

\end{document}